\documentclass[letterpaper, 10 pt, conference]{ieeeconf}  

\IEEEoverridecommandlockouts                                                                                        
\usepackage{graphics} \usepackage[dvips]{graphicx}
\usepackage{epsfig} 
\usepackage[outdir=./figures/]{epstopdf}

\usepackage{amsmath} \usepackage{amssymb}  \usepackage{multirow} 

\usepackage{caption}
\usepackage{subcaption}
\usepackage{wrapfig}

\usepackage[breaklinks]{hyperref} \hypersetup{
    pdftitle={},
    pdfsubject={2017},
    pdfauthor={David Estevez, Juan G. Victores},
    pdfkeywords={robotics} {ironing},
    colorlinks,
    citecolor=black,
    filecolor=black,
    linkcolor=black,
    urlcolor=black,
}

\usepackage[hyphenbreaks]{breakurl}
\usepackage{hypdvips}

\title{\LARGE \bf
Robotic Ironing with 3D Perception and Force/Torque\\
Feedback in Household Environments
}

\author{David Estevez, Juan G. Victores, Raul Fernandez-Fernandez and Carlos Balaguer \thanks{All of the authors are members of the Robotics Lab
    research group within the Department of Systems Engineering and Automation,
    Universidad Carlos III de Madrid (UC3M). {\tt\small destevez@ing.uc3m.es}}}

\newcommand{\norm}[1]{\left\lVert#1\right\rVert}
\DeclareMathOperator*{\argmin}{arg\,min}

\newcommand{\wildextended}{Wrinkleness Local Descriptor}
\newcommand{\wild}{WiLD}

\begin{document}

\maketitle
\thispagestyle{empty}
\pagestyle{empty}

\begin{abstract}
As robotic systems become more popular in household environments, the complexity of required tasks also increases. In this work we focus on a domestic chore deemed dull by a majority of the population, the task of ironing.
The presented algorithm improves on the limited number of previous works by joining 3D perception with force/torque sensing, with emphasis on finding a practical solution with a feasible implementation in a domestic setting.
Our algorithm obtains a point cloud representation of the working environment. From this point cloud, the garment is segmented and a custom \wildextended{} (\wild) is computed to determine the location of the present wrinkles. Using this descriptor, the most suitable ironing path is computed and, based on it, the manipulation algorithm performs the force-controlled ironing operation.
Experiments have been performed with a humanoid robot platform, proving that our algorithm is able to detect successfully wrinkles present in garments and iteratively reduce the wrinkleness using an unmodified iron.

\end{abstract}

\section{INTRODUCTION}

An increasing demand exists for robots to be capable of assisting people with everyday domestic chores. 
A great deal of robotic research has been focusing on aspects of tasks in kitchen \cite{beetz2011robotic}\cite{TamimAsfour2013} and even bathroom \cite{huete2012personal} scenarios.
However, certain domestic tasks, specially those related with deformable objects such as textiles (e.g. laundry, ironing, etc), remain a very big challenge for robots. The level of complexity is high for these tasks given the deformations and almost infinite possible configurations these objects can adopt.
Solutions in perception algorithms working with garments include the use of models which may fail to fit unknown garments \cite{Hu2009}, or fiducial markers which may even include large numbers of QR identifiers on a single garment \cite{6095109}. Regarding tasks that require advanced perception and manipulation, the ironing task is a particularly salient example. If performed incorrectly, it can induce even more wrinkles.

Focus groups and systematic studies have described ironing as a strongly disliked domestic chore and one that would be preferably automated \cite{stewardson2003market}. However, literature on robotic ironing is scarce, especially in the domain of real robotic platforms.
The EPSRC grant ``A Feasibility Study into Robotic Ironing'' from the past decade resulted in a number of theoretical analysis, such as a study on algorithms for folding and unfolding garments as well as contemporary robotic gripper solutions that could be of aid during the ironing task \cite{Dai2014Folding}, and a mathematical model of an ironing path derived from a protractor-and-tracing-paper tracking of a human demonstration \cite{Dai2014Trajectory}.

More recently, Kormushev et al. incorporated an actual physical robot and force feedback to the robotic ironing process \cite{Kormushev2011}. Ironing paths were extracted from user demonstrations by kinesthetic teaching, and force profiles were extracted from demonstrations via a haptic device.
Finally, Li et al. \cite{Li2016} included 3D perception in the ironing process to be able to directly treat wrinkles, while using a 6 cm thick foam to compensate not using force feedback for the actual physical robot setup. The perception system required a dark-room setup with controlled orthogonal light sources, as well as requiring an image of each garment without wrinkles (in the same exact pose) for each wrinkled garment to iron.

In this paper, a robotic ironing system that encompasses 3D perception and force/torque feedback evaluated on a full-sized humanoid robot is presented.
The algorithms and implementation present the following set of advantages with respect to predecessors in robotic ironing.

\begin{itemize}
\item Standard room light conditions can be used, without special illumination setup or control requirements.
\item The perception algorithm is garment-agnostic and model-agnostic, so information of the garment or even its category is not needed. 
\item Force-feedback is incorporated within the ironing loop, avoiding the use of foams and similar mechanisms.
\item The ironing loop with force-feedback is based on information from the 3D perception algorithm.
\end{itemize}

\begin{figure}[htbp]
    \centering
    \includegraphics[width=0.40\textwidth]{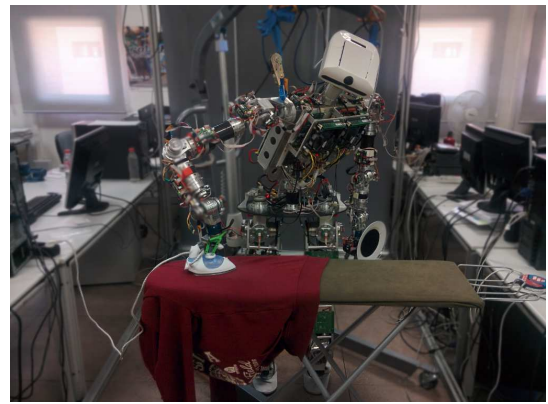}
    \caption{Experimental setup for full-sized humanoid robotic ironing with 3D perception and force/torque feedback.}
    \label{fig:setup}
\end{figure}

Figure \ref{fig:setup} depicts the experimental setup using no special lighting conditions, an arbitrary garment, and an unmodified iron and ironing board with the humanoid robot TEO. 

The complete robotic ironing algorithm will be described in section \ref{algorithm}.
The experimental setup and results will be given in section III, leading to the final section IV of conclusions.

\section{ALGORITHM}
\label{algorithm}

Our algorithm has two main components: a 3D perception-based ironing path generation and an iterative path-following ironing control. These components are designed to work sequentially, applying first the perception component to compute the most suitable ironing path, and then using the manipulation component to control the iron while performing the ironing trajectory. This sequence can be applied iteratively until no wrinkles are present in the garment patch currently placed on the ironing board. Both components are focused in the feasibility of its implementation in a domestic setting. For that purpose, the perception component only requires an RGB-D sensor, and the manipulation component works with an unmodified iron.

\subsection{3D perception-based ironing path generation }

Our perception component consists of three main stages. In the first stage, the working scene is scanned with the KinectFusion \cite{Newcombe2011} algorithm to obtain a high-resolution 3D representation of the garment and the ironing board.  The garment is then segmented from the ironing board, and a custom descriptor is computed in the second stage. Finally, in the last stage, the computed descriptor is used to find the wrinkled areas and to determine the most suitable ironing path.

The first stage starts by obtaining a 3D representation of the working scene. As RGB-D sensors typically do not have enough resolution on a single depth frame to determine precisely the presence and location of wrinkles, the KinectFusion algorithm is used to integrate data from different perspectives. The resulting 3D mesh is processed and the ironing board plane is located using RANSAC and the following prior about its location and orientation: the plane with a higher z component that has a normal aligned with the global z axis is selected. The ironing board plane contains points from both the ironing board top surface and the garment to be ironed. It is assumed that the points are linearly separable based on color, fulfilling the following requirement for two sets of points $X_0$ and $X_1$:

\begin{subequations}
\begin{align}
        \sum_{i=1}^n \omega_i x_i > k \qquad \forall x \in X_0 \\
        \sum_{i=1}^n \omega_i x_i < k \qquad \forall x \in X_1
\end{align}
\end{subequations}

Where $\omega_1, \omega_2, \dots, \omega_n$, are weights and $k$ is a real number.
Therefore, we can partition the set of $n$ points contained in the ironing board plane into $k = k_G + k_B$ different clusters, $k_G$ of them belonging to the garment and  $k_B$ belonging to the ironing board. If uniform color and a Lambertian reflectance model are assumed in both the garment and the ironing board surface, the total number of clusters to consider is $k = 1 + 1 = 2$. To improve separability we convert the color from the RGB space to the HSV space. We can then iteratively minimize the within-cluster sum of squares using the following expression for $k$ clusters:
\begin{equation}
\argmin_S \sum_{i=1}^{k} \sum_{\vec{x} \in S_i} \norm{\vec{x}-\mu_i}^2
\end{equation}

Where $k$ is the number of clusters, $S_i$ is the $i$-th cluster, $\vec{x}$ is the feature vector, and $\mu_i$ is the mean of the points in $S_i$. The feature vector includes both the spatial location of the point and the HSV color components:

\begin{equation}
\vec{x} = \left( x, y, z, h, s, v \right)
\end{equation}

\begin{figure*}[htbp]
	\centering
    \begin{subfigure}[l]{0.42\textwidth}
	    \centering
    	\includegraphics[width =\textwidth]
    	{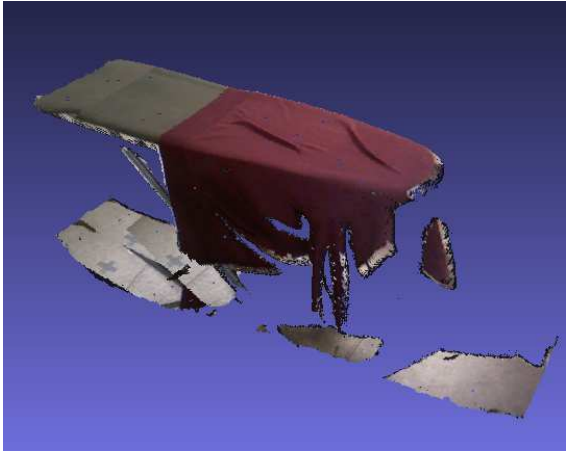}
    	\caption{Colored 3D point cloud obtained with KinectFusion}
    	~
	\end{subfigure}
	~
    \begin{subfigure}[r]{0.47\textwidth}
	    \vspace{-2ex}
	    \centering
    	\includegraphics[width =\textwidth]
    	{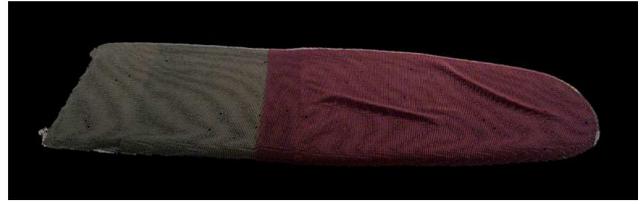}
    	\caption{Ironing board plane segmentation result}
    	\includegraphics[width =\textwidth]
    	{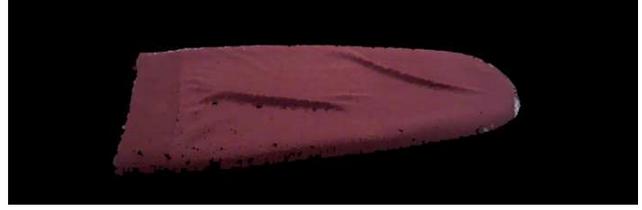}
    	\caption{Segmented garment}
	\end{subfigure} 
    \caption{3D analysis for the garment segmentation stage.}
    \label{fig:segmentation}
\end{figure*}

Once the clusters are obtained, the garment cluster is labeled accordingly based on the spatial location of its centroid. This cluster is further cleaned using Euclidean clustering to discard smaller sub-clusters within the garment cluster. Figure \ref{fig:segmentation} depicts the steps of this segmentation process.

The next stage computes a custom \wildextended{} (\wild) that is used in the last stage to locate the wrinkles. The \wild{} descriptor is computed using the following equation:

\begin{equation}
\textrm{\wild}(\vec{i}) = \frac{1}{k} \cdot \sum_{\vec{j} \in K}{\vec{n}_{\vec{i}} \cdot \vec{n}_{\vec{j}}}
\end{equation}

Where $\bar{n}_{\vec{i}}$ denotes the normal vector computed at point $\vec{i}$, $\vec{i}$ is the point for which the descriptor is being computed and $K$ is the set of $k$ neighboring points within a radius $r$. This descriptor is bounded between 0 and 1, with values closer to 1 representing planar regions and values closer to 0 corresponding to sharp edges. Wrinkles are regions not as sharp as edges so for them the descriptor values lie in intermediate values, which depend on the abruptness of the wrinkle. The different values obtained for this descriptor are projected onto a 2D image, using the garment plane as the projection plane, and z-buffering for managing possible occlusions. A segmentation mask is also generated, assigning values of 1 where a garment point was projected and 0 where no point was projected.

This \wild{} 2D image and the corresponding mask are the input for the last stage. This stage starts by performing a binary closing over the segmentation mask for filling possible artifacts due to the projection. The erosion applied during the binary closing is larger than the dilation. The reason behind this operation is to remove part of the border, since it will be always curved to some extent due to the shape of the ironing board and the effect of the gravity pulling down on the garment. The contour of the resulting mask is extracted and stored. The \wild{} image is then normalized  and thresholded to obtain the wrinkled regions:

\begin{equation}
wrinkle(\vec{x}) = \begin{cases}
    1, & \text{if } L_t < \textrm{\wild}(\vec{x}) < H_t\\
    0,              & \text{otherwise}
\end{cases}
\end{equation}

Where $L_t$ and $H_t$ are the lower and higher thresholds.

The Radius-Based Surface Descriptor (RSD) \cite{Marton2011} is an alternative descriptor that encodes the radial relationship of the point and its neighborhood, describing the curvature of the surface at a given point. 

To obtain the ironing path, the binary region is skeletonized using the Zhang-Suen algorithm. The pixels from this skeleton are converted into a graph, and the start and end points of the ironing path are selected according to the following criteria:

\begin{itemize}
\item Start point: leaf node of the graph that is closest to the garment contour.
\item Ending point: leaf node of the graph that is furthest from the garment contour.
\end{itemize}

\begin{figure}[t]
    \centering
    \includegraphics[width=0.44\textwidth]{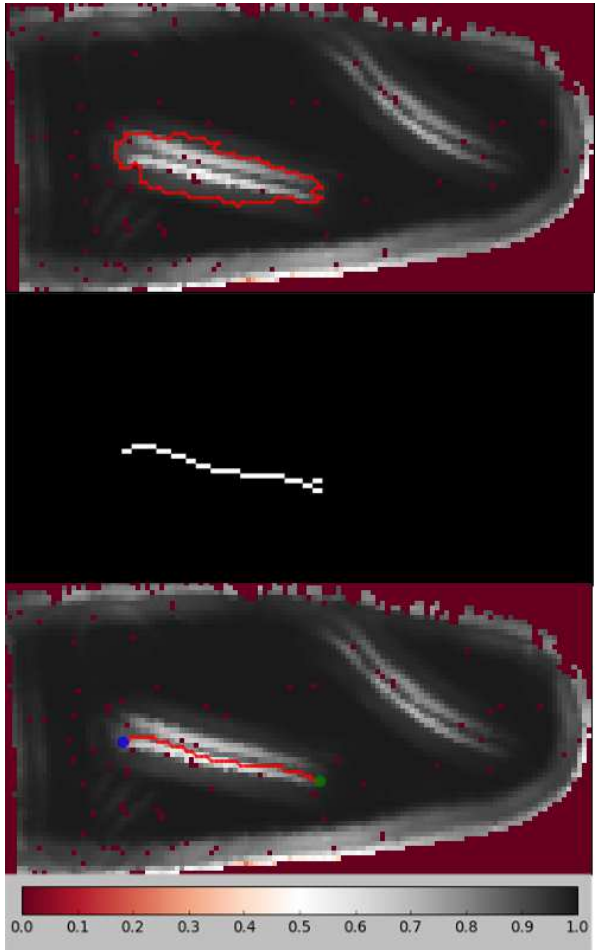}
    \caption{Ironing path extraction from \wild{} image. Images correspond (from top to bottom) to wrinkle region selection, skeletonization of selected wrinkle region and ironing path. The image colors correspond to the value of the \wild{} descriptor for each point. The start and end points of the ironing path are marked with blue and green points, respectively.}
    \label{fig:wild}
\end{figure}

Once the start and ending points are selected, the path is found by applying a depth-first search on the skeleton graph.
Figure \ref{fig:wild} depicts this process.

\subsection{Iterative path-following ironing control}

Manipulation for ironing is performed by an iterative force-controlled path-following algorithm that uses a set of $w=\{w^{(0)}, ..., w^{(N)}\}$ visually-obtained waypoints as references. Each waypoint $w^{(i)}$ is described by its Cartesian components $(w^{(i)}_{x}, w^{(i)}_{y}, w^{(i)}_{z})$.

An initial set of waypoints $p=\{p^{(0)}, ..., p^{(N)}\}$ is fixed, aimed at moving the iron to an end position over the garment with the iron perpendicular to the surface of the ironing board. The end position is at a fixed height, directly on top of the first waypoint given by the perception system. This is achieved by replacing the components ${p^{(N)}_{x}}$ and ${p^{(N)}_{y}}$ with ${w^{(0)}_{x}}$ and ${w^{(0)}_{y}}$. 

A vertical descent stage is performed until contact with the ironing board is detected. The descent is either in small commanded joint increments $\Delta{q_{cmd}}$ given a desired Cartesian increment vector {$\Delta{x_{d}}=(0,0,-{x}_{dz})$} using the Levenberg-Marquardt algorithm \cite{marquardt1963algorithm} to perform inverse kinematics, or in a velocity control loop $\dot{q}_{cmd} = J^{\dagger}_{A}(q)\cdot\dot{x}_{d}$, where $\dot{q}_{cmd}$ is the commanded joint velocity vector, $J^{\dagger}_{A}(q)$ is a Moore-Penrose pseudoinverse that uses singular value decomposition based on householders rotations with updated joint positions $q$ at each iteration, and $\dot{x}_{d}=(0,0,-\dot{x}_{dz})$ is the desired Cartesian velocity vector.
The iron descent stage ceases when the force/torque sensor measurements reach above a $f_d=((f_{dx},f_{dy},f_{dz}),(t_{dx},t_{dy},t_{dz}))$ threshold, which indicates contact.

The ironing path $w=\{w^{(0)}, ..., w^{(N)}\}$, received from the vision system, must be followed to iron the detected wrinkles.
The iterative force-controlled path-following algorithm is mandated by the control vector $\vec{c}$ at each time step $(k)$:

\begin{equation}
\vec{c}^{(k)} =
\begin{cases}
    c_{x}^{(k)} = c_{t} \cdot cos( atan2( (w^{(i)}_{y} - y^{(k)}) , (w^{(i)}_{x} - x^{(k)}) ) )\\
    c_{y}^{(k)} = c_{t} \cdot sin( atan2( (w^{(i)}_{y} - y^{(k)}) , (w^{(i)}_{x} - x^{(k)}) ) )\\
    c_{z}^{(k)} = c^{(k-1)}_{z} + K_f \cdot (f_{dz} - f^{(k)}_{z})
\end{cases}
\end{equation}

Where  $(c_{x}^{(k)}, c_{y}^{(k)}, c_{z}^{(k)})$ are the Cartesian components of $\vec{c}^{(k)}$, $c_{t}$ is a user-defined constant corresponds to the desired tangential component of $\vec{c}$, $x^{(k)}$ and $y^{(k)}$ are Cartesian components of the current position, and $K_f$ is the proportional controller gain on the vertical axis. The control vector $\vec{c}$ may be used to as $\Delta{x_{d}}$ to compute a small commanded joint increment $\Delta{q_{cmd}}$ through inverse kinematics as previously, or used in the differential form $\dot{x}_{d}$ to compute $\dot{q}_{cmd}$ through its premultiplication of $J^{\dagger}_{A}(q)$.

Finally, the vertical ascent stage is performed analogous to the descent using {$\Delta{x_{d}}=(0,0,+{x}_{dz})$} or {$\dot{x}_{d}=(0,0,+\dot{x}_{dz})$} for a certain amount of distance or time. 
\section{EXPERIMENTS}
\label{experiments}

For the validation of the proposed algorithm, a series of experiments were performed on a robotic platform. This section describes the experimental setup in which the experiments are framed, the experiments, and the results obtained from the experiments. The algorithm has been open sourced and is available online\footnote{\url{https://github.com/roboticslab-uc3m/textiles}}.

\subsection{Experimental Setup}
The experimental setup is set to emulate the situation of a robot deployed in a real household environment. For this purpose, a real ironing table was setup with real garments, and a common unmodified iron was used.

The robotic platform used for the experiments was TEO, the humanoid robot from the Robotics Lab of Universidad Carlos III de Madrid \cite{martinez2012teo}. Since grasping was not the main purpose of this work, the robot hand was removed, and the iron was installed directly in the right arm using custom 3D printed parts. This robot has a JR3 force/torque sensor in each wrist.  For 3D perception the robot head is equipped with an ASUS Xtion PRO LIVE RGB-D sensor. The pan and tilt degrees of freedom of the robot's neck were used to move the camera while scanning the scene with the KinectFusion algorithm, to generate the disparity required by the algorithm. We used the PCL implementation\footnote{\url{http://pointclouds.org/documentation/tutorials/using\_kinfu\_large\_scale.php}} of KinectFusion for our experiments.

After the scan, the 3D scene is introduced in our perception algorithm and the ironing path points are obtained. Points are converted from the image frame to the robot frame using the corresponding homogeneous transformation matrices. The robot starts performing a preprogrammed approximation trajectory to avoid collisions with the ironing board, and proceeds to descend until the iron contacts the garment. This is detected by setting the $f_d$ force/torque threshold,
which was hand-crafted compensating the deviation between the vertical axis of the sensor and the normal opposition forces generated by the ironing board.

\subsection{Experiments}
For evaluation of the algorithm presented in this paper, two different sets of experiments were designed: one for the perception algorithm, and other for the complete algorithm.
The values of the different parameters used for the two sets of experiments were the following ones: RANSAC threshold = 0.02, normal estimation radius = 0.02, \wild{} neighborhood radius = 0.03, \wild{} upper threshold = 0.95, \wild{} lower threshold = 0.4, erosion structuring element size = 11.

The first set of experiments evaluates the accuracy of the perception algorithm to detect wrinkles. 10 trials were performed with different garments, 5 with 1 wrinkle and 5 with 2 wrinkles. Only the perception component was run for these experiments, performing only the 3D scan of the garment and the wrinkled region extraction. Our algorithm is compared to RSD using the Jaccard Similarity Index (JSI) as a metric:

\begin{equation}
JSI(A, B) = \frac{\left| A \cap B\right|}{\left| A \cup B\right|}
\end{equation}

\wild{} and RSD extracted regions are compared against a hand-labeled ground truth of the scan using JSI. Computation times for both descriptors were also computed with a GNU/Linux computer equipped with a Intel(R) Core(TM) i7-4790 CPU @ 3.60GHz processor and NVidia GeForce GTX 960 graphics card using PCL-1.7.
Table \ref{table:perception_results} shows the results of this set of experiments.
On average, \wild{} presents a JSI over 25\% better than the RSD descriptor, while being more than 40\% faster. The assumption regarding precision in terms of JSI is that \wild{} has been developed for wrinkles and does not attempt to be as general as RSD. On the other hand, the simple and multi-threaded implementation of \wild{} enables much faster processing.

\begin{table*}[ht]
\vspace{1em}
\caption{Perception Algorithm: Results}
\label{table:perception_results}
\centering
\begin{tabular}{|c|r||c|c|c|c|c|c|c|c|c|c||c|}
\hline
\multicolumn{2}{|c||}{Experiment}      & \#1 & \#2 & \#3 & \#4 & \#5 & \#6 & \#7 & \#8 & \#9 & \#10 & Mean \\ \hline \hline
\multirow{2}{*}{RSD}  & JSI (\%) & 36.67   & 32.51   & 44.90   & 30.75   & 41.12   & 42.27   & 48.84   & 38.30   & 31.07   & 34.60    & 38.10   \\ \cline{2-13} 
                      & Time (s)      & 52.1271  & 50.2347   & 52.4690   &  52.3757  & 53.2446   & 50.3053   & 49.7287   & 50.4271   & 49.6802   & 51.6080    & 51.2200   \\ \hline \hline
\multirow{2}{*}{\wild{}} & JSI (\%) & 68.41 & 60.98 & 67.39 & 59.65 & 61.94 & 68.64 &  64.02 & 65.56 & 67.03 &  60.64    & 64.43   \\ \cline{2-13} 
                      & Time (s)      & 36.0774   & 34.5110   & 36.6981   & 36.6254   & 38.0608   & 34.8852   & 34.888   & 35.3923   & 34.2363   & 35.3987    & 35.6770   \\ \hline
\end{tabular}
\end{table*}

The second set of experiments evaluates the performance of the whole algorithm, including both the perception component and the manipulation component.
The values of the $f_d$ parameter used are ((0,0,-200),(0,25,0)) sensor Internal Units (IU).
For each experiment, a garment is manually placed on the ironing board. To ensure a homogeneous initial state for all the experiments, the garment is laid mostly flat and two wrinkles are manually generated randomly over the garment. This ensures all the trials start with the same number of wrinkles. The algorithm is performed iteratively, and the number of iterations and elapsed times are recorded for each iteration and trial. The results for the 5 experiments are shown in table \ref{table:algorithm_results}. While intermediate wrinkleness vary, convergence to zero wrinkles on only 2 iterations proves to be a very satisfactory result.

\begin{table*}[ht]
\vspace{1em}
\caption{Ironing Algorithm: Results}
\label{table:algorithm_results}
\centering
\begin{tabular}{|r||c|c|c|c|c|c|c|c|c|c|}
\hline
Experiment       & \#1 & \#2 & \#3 & \#4 & \#5     \\ \hline \hline
Number of iterations & 2   & 2   & 2   & 2   & 2   \\ \hline
Avg. time (s) / iteration & 71.827   & 67.803   & 72.561   & 75.684   & 63.684   \\ \hline
\end{tabular}
\end{table*}

As a metric for the progress towards the goal of a completely ironed garment, a measure of the total wrinkleness of the presented garment is computed through the following equation:

\begin{equation}\label{eq:cool}
\textrm{wrinkleness} = \frac{\sum_{\vec{x} \in G} wrinkle(\vec{x})} {\left|{G}\right|}
\end{equation}

Where $G$ is the set of points belonging to the garment and $\left|{G}\right|$ represents the cardinality of that set.

\begin{figure}[htbp]
    \centering
    \includegraphics[width=0.5\textwidth]{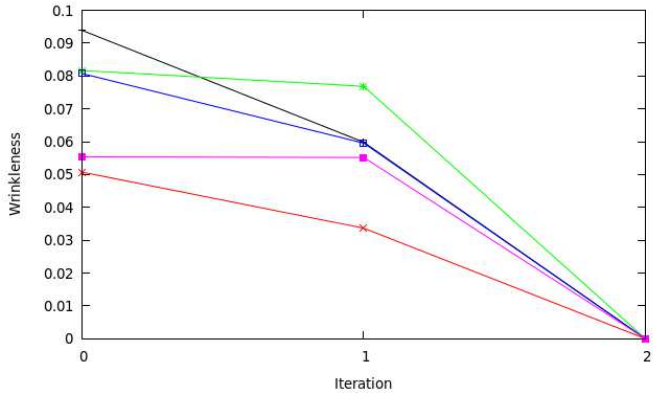}
    \caption{Wrinkleness on each experiment trial vs iterations. Iteration 0 corresponds to the initial wrinkleness. Zero wrinkleness corresponds to a situation where no wrinkles between the \wild{} thresholds were found.}
    \label{fig:graph}
\end{figure}

This value of wrinkleness is tracked for each iteration and trial, and shown in Figure \ref{fig:graph}.

\begin{figure}[htbp]
    \centering
    \includegraphics[width=0.44\textwidth]{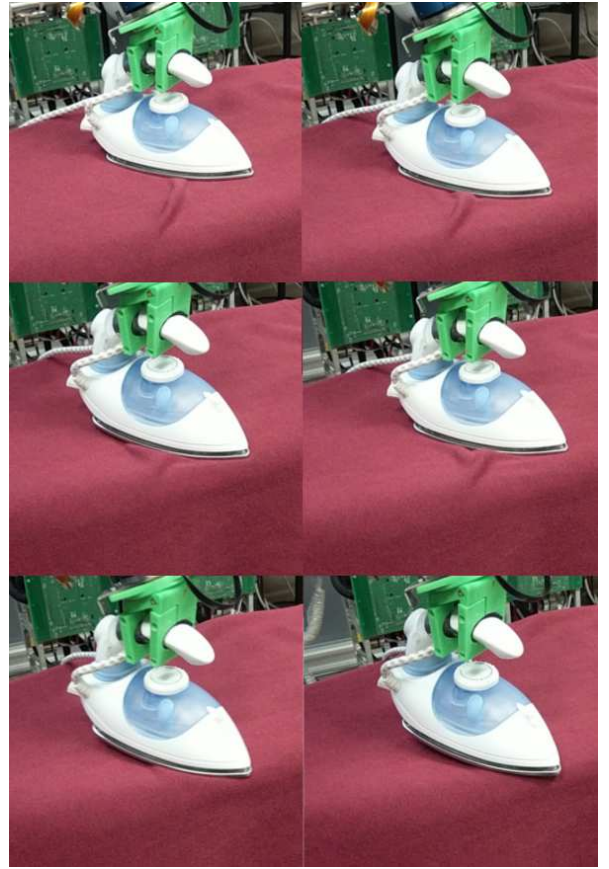}
    \caption{Ironing operation being performed during an experiment.}
    \label{fig:experiments}
\end{figure}

Figure \ref{fig:experiments} shows a sequence of still frames describing one of the ironing operations performed during the experiments. A video of the algorithm, including a subset of the performed experiments, has been submitted as complementary material.

\addtolength{\textheight}{-4cm}   

\section{CONCLUSIONS AND FUTURE WORK}
\label{conclusions}
In this work we present a robotic ironing method for a humanoid robot using unmodified human tools in an unmodified domestic environment. This work is focused on providing a practical method that has a feasible implementation in a domestic setting. For that purpose it proposes a robotic ironing system that encompasses 3D perception and force/torque feedback. Garment data is obtain by applying our segmentation stage to a colorful 3D scan of the working environment. A custom \wild{} descriptor is computed to detect wrinkled regions. Once these regions have been computed, the most suitable ironing path is estimated. The ironing path is used in conjunction with an iterative path-following ironing control to perform the ironing operation over the actual garment. This system has been evaluated on a full-sized humanoid robot. Results show that this system is capable of performing successfully ironing operations over simple garments, and it is very promising to be included in a complete ironing pipeline.

Due to the great diversity of garment shapes, textures, materials and decorative elements, ironing is a challenging task to automate. Our work, as a first approach to a practical robot ironing process, is focused in ironing simple garments. As future work we propose to extend the proposed algorithm with detection of different elements present in garments, such as buttons, zippers or other decorative elements. These elements are typically to be avoided when ironing. Different garment parts, such as collars, cuffs or pockets require different ironing approaches depending on their size and location, and their detection would benefit the proposed method. Finally, a more relaxed color uniformity constraint would allow this algorithm to work with garments with more than one colors, or even decorative patterns.

Regarding manipulation, position and velocity control have been tested with force/torque feedback. While velocity control reduces the jerk of the movements, it must be bounded to avoid excessive joint space velocities when the determinant of the Jacobian is near zero. There is obvious room for testing torque control using the transposed Jacobian matrix $J^T$ which may enable enhancements through active compliance.
In addition, the current approach could be extended by including the remaining steps of the ironing process to have a fully automated process. One of these steps is the manipulation of the garment for its placement in the ironing board. Once the current garment patch is correctly ironed, this garment has to be removed and, if necessary, placed again in the next configuration.

\section*{ACKNOWLEDGMENT}

This work was supported by RoboCity2030-III-CM project (S2013/MIT-2748), funded by Programas de Actividades I+D in Comunidad de Madrid and EU and by a FPU grant funded by Ministerio de Educación, Cultura y Deporte.
It was also supported by the anonymous donor of a red hoodie used in our initial trials.
We gratefully acknowledge the support of NVIDIA Corporation with the donation of the NVIDIA Titan X GPU used for this research.

\bibliographystyle{IEEEtran}

\bibliography{2017-iros-textiles.bib}

\end{document}